\documentclass{article}

\usepackage[preprint, nonatbib]{neurips_2021}

\usepackage[utf8]{inputenc} 
\usepackage[T1]{fontenc}    
\usepackage{url}            
\usepackage{booktabs}       
\usepackage{amsfonts}       
\usepackage{nicefrac}       
\usepackage{microtype}      
\usepackage{xcolor}         
\definecolor{citecolor}{HTML}{0071bc}
\usepackage[pagebackref=false,breaklinks=true,         colorlinks,citecolor=citecolor,bookmarks=false]{hyperref}

\usepackage{amsfonts}
\usepackage{amsmath}
\usepackage{amsthm}
\usepackage{amssymb}
\usepackage{mathtools}
\usepackage{qcircuit}
\usepackage{braket}
\usepackage[english]{babel}
\usepackage{algorithm}
\usepackage{dsfont}
\usepackage{color} 
\usepackage[framemethod=TikZ]{mdframed}
\usepackage{algpseudocode}
\usepackage{braket}
\usepackage{etoolbox}
\apptocmd{\sloppy}{\hbadness 10000\relax}{}{}

\usepackage{graphicx}
\usepackage{subcaption}

\usepackage{float}
\usepackage{wrapfig}
\usepackage{tabularx}

\usepackage{tocloft}

\definecolor{ForestGreen}{RGB}{34, 139, 34}
\definecolor{DeepSkyBlue}{RGB}{0, 191,255}
\definecolor{Lavender}{RGB}{230, 230, 250}

\newcommand{\secref}[1]{Section \ref{#1}}

\newcommand{\figref}[1]{Figure \ref{#1}}
\renewcommand*{\eqref}[1]{Equation \ref{#1}}

\newtheorem{theorem}{Theorem}

\newtheorem{definition}[theorem]{Definition}

\newcommand{\comment}[1]{\footnote{}}

\newcommand{\IBM}{\textsf{IBM}}

\DeclareMathOperator*{\argmin}{arg\,min}
\DeclareMathOperator{\KC}{\mathcal{KC}}
\DeclareMathOperator{\MMD}{\mathrm{MMD}}

\DeclareMathOperator{\QAOA}{\textsf{QAOA}}

\definecolor{dkgreen}{rgb}{0,0.6,0}
\definecolor{gray}{rgb}{0.5,0.5,0.5}
\definecolor{mauve}{rgb}{0.58,0,0.82}

\usepackage{amsmath, amsfonts, bm}

\def\1{\bm{1}}

\DeclareMathAlphabet{\mathsfit}{\encodingdefault}{\sfdefault}{m}{sl}
\SetMathAlphabet{\mathsfit}{bold}{\encodingdefault}{\sfdefault}{bx}{n}

\title{Matching Point Sets with Quantum Circuit Learning}

\author{%
	Mohammadreza Noormandipour\footnotemark[1]\quad\quad
	Hanchen Wang\thanks{Equal contribution. Correspondence: \texttt{mrn31@cam.ac.uk}}\vspace{0.7em}\\
	University of Cambridge \quad
}

\begin{document}
\maketitle
\begin{abstract}
In this work, we propose a parameterised quantum circuit learning approach to point set matching problem. In contrast to previous annealing-based methods, we propose a quantum circuit-based framework whose parameters are optimised via descending the gradients w.r.t a kernel-based loss function. We formulate the shape matching problem into a distribution learning task; that is, to learn the distribution of the optimal transformation parameters. We show that this framework is able to find multiple optimal solutions for symmetric shapes and is more accurate, scalable and robust than the previous annealing-based method. Code, data and pre-trained weights are available at the project page:
\href{https://hansen7.github.io/qKC}{https://hansen7.github.io/qKC}

\end{abstract}

\section{Introduction}

Recent advances in Noisy Intermediate Scale Quantum (NISQ)~\cite{Preskill_2018} technologies have validated the potentials to achieve a quantum advantage (i.e., supremacy~\cite{define_speedup}) with tens of (noisy) qubits.
As a subset of these techniques, quantum machine learning (QML)~\cite{qml_nature,qml_schuld,qml_book} explores the venues to deploy machine learning algorithms on quantum systems~\cite{arute2019quantum,Zhongeabe8770,reza2020_3731158}.

The structure of QML algorithms is found to be fundamentally homogeneous with that of classical kernels~\cite{schuld2021quantum}. Furthermore, kernel-based quantum models are supposed to benefit from a more efficient training than the variational variants~\cite{schuld2021quantum} and therefore need fewer training parameters. This is of importance in hybrid quantum-classical models (i.e., quantum models with classical optimisation routines) on NISQ devices, since the circuits should be kept as shallow as possible due to existence of decoherence. On this basis, we are particularly interested in advancing QML methods with kernels. 

In this paper, we propose a parameterised quantum circuit model (PQC) with a classical and a quantum kernel-based loss to match the point sets. The solution is inferred from the output distribution of the PQC and the training is performed via classical gradient descent. 
In comparison with the prior method~\cite{golyanik2020quantum}, our method, qKC, is: 1) a gate-based quantum computation model rather than an annealing-based model; 2) empirically scaled to 3D shapes; 3) differentiable trainable; 4) with smaller generalisation error; and 5) capable of finding multiple solutions for the symmetric shapes. 

\section{Problem Statement\label{sec:ps_def}}
\vspace{-1ex}
In this section, we lay out the basic problem setting and notation for matching two point sets.

We denote the two point sets as $\mathcal{M}$ (model) and $\mathcal{S}$ (scene), where $\mathcal{M} = \{\mathbf{m}_i\}_{i=1}^N$ and $\mathcal{S} = \{\mathbf{s}_i\}_{i=1}^{N'}$. $\mathcal{M}$ is assumed to be transformed from $\mathcal{S}$ via a rigid transformation: $\mathcal{T} = [\mathbf{r}_{\mathcal{MS}}, \mathbf{t}_{\mathcal{MS}}]$, where $\mathbf{r}_{\mathcal{MS}}\in $ SO(2) or SO(3) and $\mathbf{t}_{\mathcal{MS}}\in\mathbb{R}^2$ or $\mathbb{R}^3$. The matching task objective is to minimise the summation of mean-squared error $\mathcal{L}$, given a ground true point pair correspondence between $\mathcal{M}$ and $\mathcal{S}$.

A point pair correspondence function $L_{\mathcal{MS}}$ is then defined to fetch the corresponding point(s) in $\mathcal{S}$ for each queried point $\mathbf{m}_i$ in $\mathcal{M}$: $L_{\mathcal{MS}}: \mathcal{M} \rightarrow \mathcal{S}$.
Usually, $L_{\mathcal{MS}}$ is assumed to be bijective ($N=N'$): $L_{\mathcal{MS}}(\mathbf{m}_i) = \mathbf{s}_j$, which is mostly used in prior works. 
An alternative design is to return multiple corresponding points in $\mathcal{S}$ for each query point $\mathbf{m}_i$ in $\mathcal{M}$:
$ L_{\mathcal{MS}}(\mathbf{m}_i) = \{\mathbf{s}_{j_k}\}_{k=1}^K$.
This multi-linked design is utilised in works including EM-ICP~\cite{EM-ICP}, SoftAssignment~\cite{softassign}, as well as ours.

With $L_{\mathcal{MS}}$ defined, we can directly solve the optimal transformation $\mathcal{T}^{opt}$ in a deterministic fashion:
\begin{equation}
    \left.\left(\frac{\partial}{\partial\mathcal{T}} \sum_{\mathbf{m}_i\in\mathcal{M}} ||\mathcal{T}\mathbf{m}_{i} - L_{\mathcal{M S}}\left(\mathbf{m}_{i}\right)||_2\right)\right|_{\,\,\mathcal{T}=\mathcal{T}^{opt}} = 0
\end{equation}
where $\mathcal{T}\mathbf{m}_{i}:= \mathbf{r}_{\mathcal{MS}}\cdot \mathbf{m}_i + \mathbf{t}_{\mathcal{M S}}$. The ground truth of the transformation $\mathcal{T}^{gt}$ is also computed in this fashion based on the provided rather than the predictive correspondence.
Clearly it is with no challenge if the predictive correspondence is the same as the ground truth. However, finding such correspondence mapping for a perfect matching is known as a NP-hard problem, and it becomes more challenging if the two point sets are less overlapped~\cite{predator}. Therefore, we tackle the matching problem under the context of fully-connection where the connected link is characterised by kernels~\cite{tsin2004correlation}.  

\begin{figure}[t]
\centering
    \includegraphics[width=\linewidth]{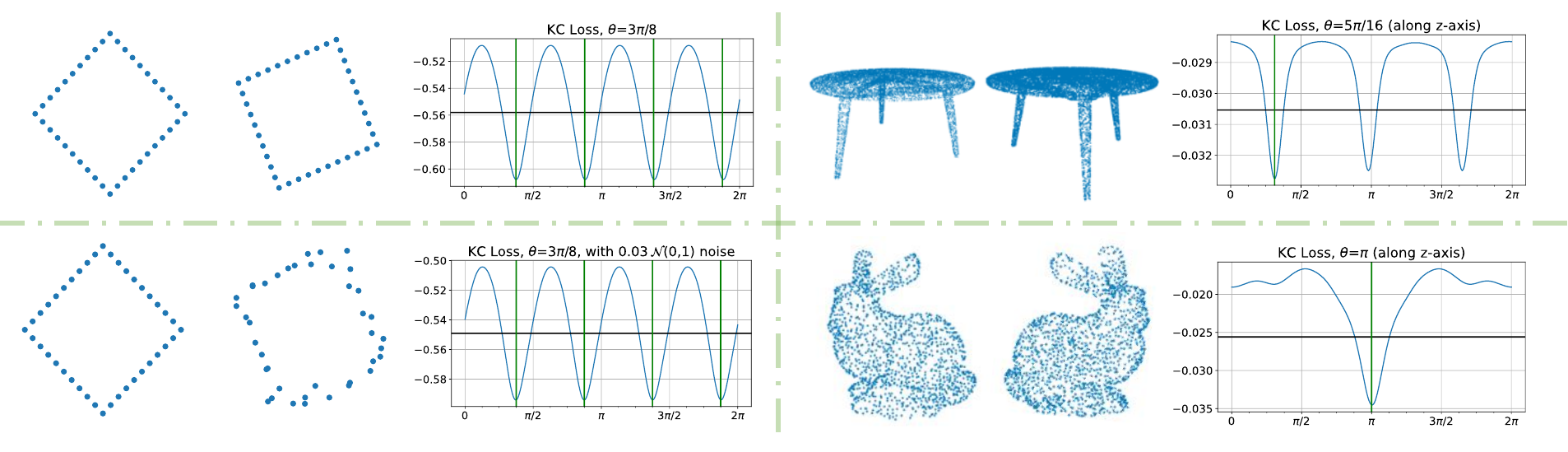}
    \caption{Diagram of the kernel correlation (KD) loss. The left column shows that KC loss is capable of finding \textit{all} the four optimal matching angles for a symmetric 2D square even in the presence of noise (bottom-left). The right column shows the effectiveness of KC loss for more complicated 3D shapes, namely a symmetric table (top-right) and an asymmetric bunny (bottom-right).\label{fig:diag_kc}}
    \vspace{-2ex}
\end{figure}

\section{Kernel Correlation\label{sec:kc_sec}}
In this section, we introduce how prior methods~\cite{tsin2004correlation,Shen_2018_CVPR} match two point sets with a kernel correlation (KC) loss. We provide a diagram of the KC loss for 2D/3D, (a)symmetric shapes in~\figref{fig:diag_kc}.

Kernel Correlation (KC)~\cite{tsin2004correlation} extends the correlation techniques to point set matching. It is a measure of affinity as well as a function of the entropy. KC between two points, $\mathbf{x}_i$ and $\mathbf{x}_j$, is defined as:
\begin{gather}\label{eq:og_kc}
    \KC(\mathbf{x}_i, \mathbf{x}_j) = \int_{\mathbb{R}^k} \kappa(\mathbf{x}, \mathbf{x}_i)\kappa(\mathbf{x}, \mathbf{x}_j)\mathrm{d}\mathbf{x}
\end{gather}
where $\kappa(\cdot, \cdot)$ is a kernel such as Gaussian and quantum. 
The homologic Gaussian KC is written as:
\begin{gather}\label{eq:kc_pointpair}
    \KC_G(\mathbf{x}_i, \mathbf{x}_j) = \alpha\exp(-||\mathbf{x}_i-\mathbf{x}_j||_2^2/\sigma^2)
\end{gather}
where $\alpha$ and $\sigma$ are constants. 
The KC between two point sets, $\mathcal{M}$ and $\mathcal{S}$, is defined as:
\begin{gather}\label{eq:kc_pointset}
    \KC(\mathcal{M}, \mathcal{S}) = \sum_{\mathbf{s} \in \mathcal{S}} \sum_{\mathbf{m} \in \mathcal{M}} \KC(\mathcal{T} \mathbf{m}, \mathbf{s}) = \mathbb{E}_{\,\mathbf{s}\sim\mathcal{S}\,\mathbf{m}\sim\mathcal{M}}[\KC(\mathbf{m}, \mathbf{s})] * \|\mathcal{M}\| * \|\mathcal{S}\|
\end{gather}
where $\|\cdot\|$ is the set cardinality.
If the two point sets are closely aligned, the KC is large. 
Therefore, the optimal transformation $\mathcal{T}^{opt}$ is solved via finding the \textit{minima} of the negative KC value:
\begin{gather}\label{eq:multi-linked-loss}
    \mathcal{T}^{opt} = \argmin\mathcal{L}_{\KC},\qquad\text{where }\mathcal{L}_{\KC} =  -\KC(\mathcal{T}\circ\mathcal{M},\,\,\mathcal{S})
\end{gather}
Notice that in~\eqref{eq:multi-linked-loss} each transformed model point $\mathbf{m}$ is interacting with all the scene points. We call~\eqref{eq:multi-linked-loss} a fully-linked registration cost function. This is in contrast to the methods like ICP~\cite{BeslM92, 132043, zhang1994iterative} and prior quantum method~\cite{golyanik2020quantum}, where each model point is connected to a subset of scene points. It is clear that the objective defined in~\eqref{eq:multi-linked-loss} satisfies the minimum requirement for a registration algorithm. That is, $\mathcal{T}^{opt}$ corresponds to one of the global minima of the cost. Same as previous work~\cite{golyanik2020quantum}, we assume the translation between two point sets is resolved by aligning the centres of mass, in the following sections, we focus on solving the optimal rotations with QCL.

\section{Quantum Circuit Learning\label{sec:qcl}}
In this section, we describe how we tackle the problem in the context of Quantum Circuit Learning (QCL)~\cite{farhi2018classification,PhysRevA.98.032309}.
In~\secref{subsec:qcbm}, we summarise how previous work utilises Born Machine circuit, a subset of PQCs for distribution learning.
In~\secref{subsec:loss}, we define a new loss and its gradient, bridging the distribution learning and the minimisation of kernel correlation loss. 
In~\secref{subsec:qkernel}, we provide theorems and details for implementing a quantum kernel, which can offer potential advantages. It is worth noting that our method is \textit{fundamentally} different from the prior attempt~\cite{golyanik2020quantum} and its follow-ups~\cite{benkner2020adiabatic,benkner2021q,birdal2021quantum} regarding the model/approach of quantum computation at work. 

In gate-based quantum computation models, quantum circuits are composed of qubits and logic (parameterised) quantum gates. From computational point of view, a qubit is a mathematical object composed of a linear combination of two basis states that the information is encoded in: $|\psi\rangle=\alpha|0\rangle+\beta|1\rangle$, where $\alpha, \beta\in\mathbb{C}$, $\|\alpha\|+\|\beta\|=1$ and $|0\rangle$ and $|1\rangle$ are the basis states in Dirac notation. The quantum gates are unitary (norm-preserving) matrices that act on a normalised initial quantum state and cause a unitary evolution of it. The choice of the unitary gates is done in such a way that the final evolved quantum state provides the solution of a specific problem upon measurements. For a more comprehensive introduction, we recommend~\cite{nielsen2000quantum,PhysRevA.98.032309}.

\subsection{Quantum Circuit and Ising Born Machine\label{subsec:qcbm}}
\begin{wrapfigure}[8]{r}{0.45\textwidth}
\vspace{-24pt}
\begin{align}\scriptsize
    \nonumber
    \Qcircuit @C=1.0em @R=0.8em {
    \lstick{\ket{0}}    & \gate{H}  & \multigate{3}{U_z(\boldsymbol\alpha)} & \gate{U^1_f(\Gamma_1, \Delta_1, \Sigma_1)}  & \meter &\cw & \rstick{x_1} \\
    \lstick{\ket{0}}    & \gate{H}  & \ghost{U_z(\boldsymbol\alpha)}        & \gate{U^2_f(\Gamma_2, \Delta_2, \Sigma_2)}  & \meter &\cw & \rstick{x_2} \\
    \cdots              &           &           & \cdots                    & 
    \cdots              &           &  \\
    \lstick{\ket{0}}    & \gate{H}  & \ghost{U_z(\boldsymbol\alpha)}        & \gate{U^n_f(\Gamma_n, \Delta_n, \Sigma_n)}  & \meter &\cw & \rstick{x_n} 
    }
\end{align}
\vspace{-8pt}
\renewcommand{\captionlabelfont}{\footnotesize}
\caption{\footnotesize Schematic of a PQC, from~\cite{coyle2020born}\label{fig:qibm-overview}}
\end{wrapfigure}
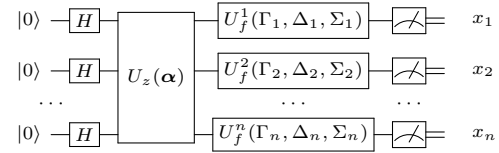
A generic PQC model (in~\figref{fig:qibm-overview}) consists of three stages: state preparation, evolution and measurements. During preparation, $n$ qubits are initialised in $\ket{0}$ state and then are acted upon by Hadamard gates to produce an unbiased super-position of possible states with equal probabilities: 
\begin{gather}
\boldsymbol{H}^{\otimes n}\ket{0}^{\otimes n} = \ket{+}^{\otimes n}
\end{gather}
It is then followed by a unitary evolution by $m$ operators, each acting on the qubits in the set $S_j$:
\begin{gather}\label{eq:evolution_unitary}
\boldsymbol{U}_z(\boldsymbol\alpha) \coloneqq \prod_{j=1}^m U_z (\alpha_j, S_j) \coloneqq \prod_{j=1}^m \exp\left(\mathrm{i} \alpha_j \bigotimes_{k \in S_j} Z_k \right)
\end{gather}
The observation is made via the measurement unitary $U_{f}(\cdot, \cdot, \cdot)$, a set of single qubit gates:
\begin{gather}\label{eq:finalmeasurementgate}
U_f\left( \mathbf{\Gamma}, \mathbf{\Delta}, \mathbf{\Sigma} \right) \coloneqq \exp\left(\mathrm{i}\sum\limits_{k=1}^n \Gamma_k X_k + \Delta_k Y_k +\Sigma_k Z_k\right)\qquad 
\end{gather}
where $X_k, Y_k, Z_k$ are the canonical Pauli operators~\cite{nielsen2000quantum} acting on the $k$-th qubit, $\Gamma_k, \Delta_k, \Sigma_k$ are parameters that can be specifically chosen to realise different circuit families incl. instantaneous quantum polynomial time (IQP)~\cite{iqp,iqp_supremacy} and quantum approximate optimisation algorithm (QAOA)~\cite{qaoa,qaoa_supremacy}. Essentially here learning with quantum circuits is to optimise the parameters $\boldsymbol\alpha$ in the unitary evolution given a particular task and a specific circuit design (e.g., interconnect). 

A Born Machine~\cite{Cheng_2018, Diff_QBM} is a quantum-classical hybrid generative model for distribution learning. Concretely, it utilises a classical optimisation tactic (i.e., gradient descent) to alter the parameters $\boldsymbol\alpha$ of a quantum system, thus the output binary vector $\boldsymbol x$ of quantum system (circuit) is able to produce samples according to the target distribution. According to the Born's Rule~\cite{Preskill_2018}, the output probability distribution of a quantum circuit (parameterised by $\boldsymbol\theta = \{\boldsymbol{\alpha, \Gamma, \Delta, \Sigma}\}$) can be written as:
\begin{gather}\label{eq:output_dist}
    p_{\boldsymbol\theta}(\mathbf{x}) = \left|\bra{\mathbf{x}}\boldsymbol U_f \left( \mathbf{\Gamma}, \mathbf{\Delta}, \mathbf{\Sigma} \right) \boldsymbol U_z(\boldsymbol\alpha)\boldsymbol{H}^{\otimes n}\ket{0}^{\otimes n}\right|^2
\end{gather}
where the raw output of the $n$-qubit circuit, $\mathbf{x}=[x_1, x_2, .., x_n]\in\{0,1\}^n$, is a $n$-dimensional binary vector. Usually, the Born Machine is restricted into an \textit{Ising} version~\cite{coyle2020born} in order to be implementable on the NISQ devices. That is, all the unitary gates act on either one or two qubits, i.e., $|S_j|\leq 2$.

On this basis, we can rewrite the unitary evolution (\ref{eq:evolution_unitary}) as an Ising Hamiltonian~\cite{bremner2016average,fujii2017commuting}:
\begin{gather}
    \boldsymbol{U}_{z}(\boldsymbol{\alpha}) = \exp{\left(\mathrm{i}\mathcal{H}_{z}\right)} \quad\Rightarrow\quad \mathcal{H}_{z} = \prod_{j=1}^{m} \alpha_{j} \bigotimes_{k \in S_{k}} Z_{i} = \sum_{i<j} J_{i j} Z_{i} Z_{j}+\sum_{k=1}^{n} b_{k} Z_{k}
\end{gather}
where $Z_{i,j,k}$ are the $z$-basis Pauli operators acting on the $\{i,j,k\}$-th qubit, $\boldsymbol\alpha = \{J_{ij}, b_k\}$ are the parameters we need to optimise w.r.t. input distribution\footnote{under some circumstances~\cite{coyle2020born}, $(\mathbf{\Gamma}, \boldsymbol{\Delta}, \mathbf{\Sigma})$ can be treated as learnable parameters as well.}. $\{J_{ij}, b_k\}$ can be viewed as the pairwise coupling and local magnetic fields, respectively. For a $n$-qubit Ising Born Machine model, the number of parameters are: $n*(n-1)/2 + n = n*(n+1)/2$. Note that the prior work~\cite{golyanik2020quantum} does not have \textit{any} parameters to be optimised, their quadratic objective is fixed and universal for all point sets.

As mentioned, the output of a $n$-qubit Ising Born Machine (IBM) is a $n$-dimensional binary vector which has a total of $2^n$ possible values. We now define the mapping from these binary output to the solution of the point matching problem. We use the same setting as previous attempt~\cite{golyanik2020quantum}, assuming the translation in the $\mathcal{T}^{\text {opt}}$ is resolved by aligning the mass centres of the two point sets, and only considering rotation along \textit{one} axis. Therefore we simply require to transform the distribution of the output binary vectors to a \textit{single} rotation angle. Specifically, we equally split the range $[0, 2\pi)$ of the rotation angle into $2^n$ number of bins, where each binary vector corresponds to a unique bin. 

For instance, if the measurement outcome of a 4-qubit IBM is $\mathbf{x} = (0, 1, 0, 1)$, the corresponding rotation angle is the median value of the 6-th bin, which is $\frac{(1\times 2^0 + 1\times 2^2)  * 2  + 1}{2^4*2} \times 2\pi = \frac{11}{16}\pi$. The resolution for the angle distribution is $2\pi/2^n$, which will grow exponentially as the number of qubits increases. However, due to the hardness of using classical simulation on quantum systems~\cite{bremner2011classical,fujii2017commuting}, in experiment sections we mainly consider 4-and 6-qubits shallow circuit architectures.

\subsection{Training Procedures\label{subsec:loss}}
We first describe how we design a kernel-based loss, to bridge the distribution learning and optimal alignment. Then we present the gradient calculation procedure to enable differentiable training.

To utilise Born Machine models for distribution learning, a differentiable loss function is usually required which measures the closeness between the model output distribution $p_{\boldsymbol{\theta}}(\mathbf{x})$ and the ground true distribution $\pi(\mathbf{y})$. The first metric to be considered is KL Divergence:
\begin{gather}
\mathcal{L}_{\mathrm{KL}}=-\sum_{\mathbf{x}} \pi(\mathbf{x}) \log \left(p_{\boldsymbol{\theta}}(\mathbf{x})\right)=-\,\mathbb{E}_{\mathbf{x} \sim \pi} \log \left(p_{\boldsymbol{\theta}}(\mathbf{x})\right)
\end{gather}
However, as noted in~\cite{bremner2011classical,coyle2020born,Diff_QBM}, calculating the gradient of $\mathcal{L}_{\mathrm{KL}}$ is \#P-hard, thus make it a suboptimal choice. So we instead use an alternative discrepancy measure called Integral Probability Metrics (IPMs)~\cite{sriperumbudur2009integral}, among which a well-known example is Maximum Mean Discrepancy (MMD)~\cite{sriperumbudur2011universality,sriperumbudur2010hilbert}\footnote{examples of IPMs incl. Wasserstein Distance, Total Variation, Stein Discrepancy and Sinkhorn Divergence.}. It is widely used in applications incl. homogeneity testing~\cite{gretton2006kernel} and independence testing~\cite{gretton2007kernel,fukumizu2007kernel}.

In information theory and probability theory, a general form of an IPM is defined as:
\begin{gather}
\gamma_{\mathcal{F}}(\mathcal{P}, \mathcal{Q})
=\sup _{\phi \in \mathcal{F}}\left|\int_{\mathcal{M}} \phi \,\,\text{d}\,\mathcal{P}-\int_{\mathcal{M}} \phi\,\,\text{d}\,\mathcal{Q}\,\right|
=\sup _{\phi \in \mathcal{F}}\left(\mathbb{E}_{\mathcal{P}}[\phi]-\mathbb{E}_{\mathcal{Q}}[\phi]\right)
\end{gather}
where $\mathcal{P}$ and $\mathcal{Q}$ are defined on the same probability space $\mathcal{M}$. $\mathcal{F}$ is a set of real bounded measurable functions on $\mathcal{M}$. $\phi(\cdot)$ defines a feature map from the input space $\mathcal{X}$ to the feature space $\mathcal{H}$ (i.e. Reproducing Kernel Hilbert Space, RKHS~\cite{aronszajn1950theory}) with a corresponding kernel function $\kappa(\cdot, \cdot)$:
\begin{gather}
\phi: \mathcal{X} \rightarrow \mathcal{H},
\qquad\quad
\kappa(\mathbf{x},\mathbf{y}) = \langle\phi(\mathbf{x}), \phi(\mathbf{y})\rangle_{\mathcal{H}},\,\,\,\forall\mathbf{x},\mathbf{y}\in\mathcal{X}
\end{gather}
$\gamma_{\mathcal{F}}(\cdot,\cdot)$ becomes MMD when $\mathcal{F}$ is confined in a unit hyper-sphere: $\mathcal{F}=\{\phi: \|\phi\|_\mathcal{H}\leq1\}$. It has been proven that MMD equates the mean embedding differences between $\mathcal{P}$ and $\mathcal{Q}$~\cite{bremner2016average}:
\begin{gather}
\gamma_{\mathrm{MMD}}(\mathcal{P}, \mathcal{Q})=\left\|\mu_{\mathcal{P}}-\mu_{\mathcal{Q}}\right\|_{\mathcal{H}} = \left\|\mathbb{E}_{\mathbf{x}\sim \mathcal{P}}[\phi(\mathbf{x})]-\mathbb{E}_{\mathbf{x}\sim \mathcal{Q}}[\phi(\mathbf{x})]\right\|_{\mathcal{H}}
\end{gather}
Using this, Born Machine~\cite{coyle2020born,Cheng_2018,Diff_QBM} defines the following MMD loss for the training:
\begin{gather}
    \mathcal{L}_{\mathrm{MMD}} 
    = \gamma_{\mathrm{MMD}}^2 
    = \mathbb{E}_{\,\mathbf{x} \sim \mathcal{P},\,\mathbf{y} \sim \mathcal{P}}[\kappa(\mathbf{x}, \mathbf{y})] 
    + \mathbb{E}_{\,\mathbf{x} \sim \mathcal{Q},\,\mathbf{y} \sim \mathcal{Q}}[\kappa(\mathbf{x}, \mathbf{y})]
    - 2\,\mathbb{E}_{\,\mathbf{x} \sim \mathcal{P},\,\mathbf{y} \sim \mathcal{Q}}[\kappa(\mathbf{x}, \mathbf{y})]
\end{gather}
By observation, we can re-write the $\mathcal{L}_{\mathrm{MMD}}$ in terms of Kernel Correlation introduced in~\eqref{eq:kc_pointset}:
\begin{gather}\label{eq:Lmmd}
    \mathcal{L}_{\mathrm{MMD}} * N^2 = 
    \mathbb{E}_{\,\mathbf{x} \sim \mathcal{P},\, \mathbf{y} \sim \mathcal{P}}[\mathcal{KC}(\mathbf{x}, \mathbf{y})] +  
    \mathbb{E}_{\,\mathbf{x} \sim \mathcal{Q},\, \mathbf{y} \sim \mathcal{Q}}[\mathcal{KC}(\mathbf{x}, \mathbf{y})] - 
   2\,\mathbb{E}_{\,\mathbf{x} \sim \mathcal{P},\,\mathbf{y} \sim \mathcal{Q}}[\mathcal{KC}(\mathbf{x}, \mathbf{y})]
\end{gather}
here we assume $\|\mathcal{P}\| = \|\mathcal{Q}\| = N$, i.e., the two sets have the same $N$ number of points. Note that the kernel $\kappa(\cdot, \cdot)$ should be \textit{bounded, measurable} and \textit{characteristic}~\cite{sriperumbudur2009integral}. These conditions are satisfied by Gaussian kernel used in~\eqref{eq:kc_pointpair}, we will later show that quantum kernels are also validated.

Notice that if $\mathbf{x}$ and $\mathbf{y}$ are sampled from the same distribution (i.e., a surface of rigid object), the $\KC$ term essentially represents the compactness of that shape, which is constant. Therefore, \eqref{eq:Lmmd} tells us minimising the $\mathcal{L}_\mathrm{MMD}$ substantially equates the diminution of the KC loss (\eqref{eq:multi-linked-loss}).

Recall that the raw output from the quantum circuit serves as the distribution of the rotation angles $p_\theta(\cdot)$, $\mathcal{P}$ and $\mathcal{Q}$ are identical. Thus we can re-write~\eqref{eq:Lmmd} in terms of $p_\theta$:
\begin{gather}\label{eq:Lmmd_angle}
    \mathcal{L}_{\mathrm{MMD}} * N^2 = - 
   2\,\mathbb{E}_{\,\mathbf{m} \sim \mathcal{M},\,\mathbf{s} \sim \mathcal{S},\,\mathbf{x}\sim p_\theta}[\mathcal{KC}(\mathcal{T}_\mathbf{x}\mathbf{m}, \mathbf{s})] + \text{constant}
\end{gather}
In~\eqref{eq:Lmmd_angle}, we have shown that using the same distribution learning framework under (Ising) Born Machine models, we can minimise the KC loss~\cite{tsin2004correlation} to align two point sets. We name our method \textbf{qKC}.  Furthermore, since the IBM solver outputs a distribution rather than a scalar/vector, qKC is able to learn \textit{multiple} optimal angles particularly for symmetric shapes (shown in~\figref{fig:training}).

It is known~\cite{coyle2020born,Diff_QBM} that the gradients of the Born Machine model can be written as\footnote{A more precise notation is to use $\mathbf{z}$ instead of $\mathbf{x}$~\cite{farhi2018classification}, since the measurement is on the $\mathbf{z}$-computational basis (i.e., electron spin around the $\hat{\mathbf{z}}$-direction): $\ket{\mathbf{z}}\bra{\mathbf{z}}$ (right after $U_{f}^{k}(\cdot, \cdot, \cdot)$ in~\figref{fig:qibm-overview}). For the sake of simplicity, here we stick to $\mathbf{x}$ to represent the measured bit string from the quantum circuit.}:
\begin{gather}\label{eq:grad}
\frac{\partial p_{\boldsymbol{\theta}}(\mathbf{x})}{\partial \boldsymbol{\theta}_{k}}=p_{\boldsymbol{\theta}_{k}}^{-}(\mathbf{x})-p_{\boldsymbol{\theta}_{k}}^{+}(\mathbf{x})
\end{gather}
\begin{wrapfigure}[5]{r}{0.45\textwidth}
\vspace{-22pt}
\begin{align}\scriptsize
    \nonumber
    \Qcircuit @C=0.6em @R=0.5em {
    & \gate{H^{\otimes n}} & \gate{U_{l:k+1}}&\gate{U_k({\theta}_k^{\pm})}&\gate{U_{k-1:1}} & \meter &\cw
    }
\end{align}
\vspace{-8pt}
\renewcommand{\captionlabelfont}{\footnotesize}
\caption{\footnotesize a Shifted PQC, from~\cite{coyle2020born,Diff_QBM}\label{fig:qibm-shift}}
\end{wrapfigure}
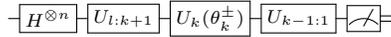
where $p_{\boldsymbol{\theta}_{k}}^{\pm}$ is the measured output distribution with a shift on the $k$-th circuit parameter $p_{\boldsymbol{\theta}_{k}}$~\cite{schuld2019evaluating}:
\begin{gather}
p_{\boldsymbol{\theta}_{k}}^{\pm}=p^{}_{\boldsymbol{\theta}_{k} \pm \pi / 2}
\end{gather}
Measuring with a parameter shifted circuit is sketched in~\figref{fig:qibm-shift}.
The derivative of $\mathcal{L}_\mathrm{MMD}$ is:
\begin{gather}
\frac{\partial \mathcal{L}_{\mathrm{MMD}}}{\partial \boldsymbol{\theta}_{k}}
= 2\,\mathbb{E}_{\,\mathbf{m} \sim \mathcal{M},\, \mathbf{s} \sim \mathcal{S},\, \mathbf{x}^{+} \sim p_{\boldsymbol{\theta}_{k}}^{+}}\left[\KC\left(\mathcal{T}_{\mathbf{x}^{+}} \mathbf{m},\mathbf{s}\right)\right] 
- 2\,\mathbb{E}_{\,\mathbf{m} \sim \mathcal{M},\, \mathbf{s} \sim \mathcal{S},\, \mathbf{x}^{-} \sim p_{\boldsymbol{\theta}_{k}}^{-}}\left[\KC\left(\mathcal{T}_{\mathbf{x}^{-}} \mathbf{m}, \mathbf{s}\right)\right]
\end{gather}
We can also write down the empirical estimation of the loss and gradient:
\begin{gather}
\begin{split}\label{eq:empirial_lmmd}
    \tilde{\mathcal{L}}_{\mathrm{MMD}} &= \tilde{\gamma}_{\mathrm{MMD}}^{2} = \frac{-2}{N^2\|\mathcal{X}\|} \sum_{\mathbf{s} \in \mathcal{S}} \sum_{\mathbf{m} \in \mathcal{M}} \sum_{\mathbf{x}\in\mathcal{X}} \KC\left(\mathcal{T}_{\mathbf{x}} \mathbf{m}, \mathbf{s}\right) \\
    \frac{\partial \mathcal{\tilde{L}}_{\MMD}}{\partial {\boldsymbol\theta}_k} &=  
    \frac{2}{N^2\|\mathcal{X}\|}\sum_{\mathbf{s} \in \mathcal{S}} \sum_{\mathbf{m} \in \mathcal{M}}\left(\sum_{\,\,\,\,\mathbf{x}\in\mathcal{X}^+} \KC(\mathcal{T}_{\mathbf{x}^+}\mathbf{m}, \mathbf{s}) - \sum_{\,\,\,\,\mathbf{x}\in\mathcal{X}^-} \KC(\mathcal{T}_{\mathbf{x}^-}\mathbf{m}, \mathbf{s})\right)
\end{split}
\end{gather}
where $\mathcal{X}$ is the number of samples from the circuit at each iteration, $\|\cdot\|$ is the cardinality, and $\|\mathcal{X}\| = \|\mathcal{X}^{+}\| = \|\mathcal{X}^{-}\|$. Intuitively the larger the sample size is, the better it can represent $p_{\boldsymbol\theta}(\mathbf{x})$ and $p_{\boldsymbol\theta^\pm}(\mathbf{x})$, as well as the loss and gradient. \cite{sriperumbudur2009integral} shows this empirical estimation on the MMD loss and gradient of qKC is unbiased and with a quadratic convergence rate:
\begin{gather}\label{eq:mmd_convergence}
\left|\tilde{\gamma}_{\mathrm{MMD}}\left(\mathcal{P}_{M}, \mathcal{Q}_{N}\right)-\gamma_{\mathrm{MMD}}(\mathcal{P}, \mathcal{Q})\right|=O_{\mathcal{P}, \mathcal{Q}}\left(\frac{1}{\sqrt{M}}+\frac{1}{\sqrt{N}}\right)\stackrel{a . s .}{\longrightarrow} 0 \,\,\,\text {as}\,\,\, M, N \rightarrow \infty
\end{gather}
where $M$ represents number of samples from distribution $\mathcal{P}$: $M=\|\mathcal{P}_M\|$. 

However, in qKC, both $\mathcal{P}$ and $\mathcal{Q}$ (or $\mathcal{M}$ and $\mathcal{S}$) represent the same underlying shape. Empirically we don't observe that sampling more points results in a faster convergence and better solution. On this basis, we further investigate how $\|\mathcal{X}\|$ (i.e., batch size) will affect the training. In experiments we find that $\|\mathcal{X}\|$ is not as critical as other parameters such as learning rate for the optimisation process. We will have more discussions on this parameter tuning in next section and appendix.

Besides~\eqref{eq:grad}, there is an alternative strategy~\cite{farhi2018classification} for the gradient estimation of Born Machine models. It requires additional measurements on the imaginary parts of the unitary via additional auxiliary qubits and Hadamard gates. Our framework by contrast is more implementation friendly.

\subsection{Quantum Kernel\label{subsec:qkernel}}

A natural extension is to realise a quantum version of the  kernel~\cite{schuld2019quantum} in~\eqref{eq:Lmmd_angle}. This might exhibit supremacy when the input is in quantum format~\cite{rebentrost2014quantum} or the feature space is classically hard to learn~\cite{havlivcek2019supervised}.
We first give a formal definition of a quantum RKHS and kernel, then propose a procedure to encode a point set with quantum kernels.

\begin{definition}[Quantum Kernel and RKHS]\label{theo:hilbert_kernel}
Let $\Phi:\mathcal{X}\rightarrow \mathcal{H}_q$ be a quantum feature map over an input set $\mathcal{X}$, which gives rise to a \emph{real} kernel $\kappa_q(\cdot, \cdot)$ and the corresponding RKHS $\mathcal{H}_q$:
\begin{gather}
\kappa_q(\mathbf{x}, \mathbf{y})
= \langle\Phi(\mathbf{x}), \Phi(\mathbf{y})\rangle_{\mathcal{H}_q} 
= |\braket{\Phi(\mathbf{x})\,|\,\Phi(\mathbf{y})}|^2 , \quad \forall \mathbf{x}, \mathbf{y} \in \mathbb{R}^d \\\vspace{7ex}
\mathcal{H}_{q}=\left\{f:\mathbb{R}^d \rightarrow \mathbb{R}\left|f(\mathbf{x})=|\langle h \left|\, \Phi(\mathbf{x})\rangle\right|^{2},\,\, \forall\, \mathbf{x} \in \mathbb{R}^d,\,\, h \in \mathcal{H}_q\right\}\right.
\end{gather}
\end{definition}
where $\Phi(\cdot)$ is realised via a parameterised quantum circuit. Note that the physical interpretation of a quantum kernel the overlap of two states in a quantum RKHS, so it is defined as the \textit{real transition probability}~\cite{schuld2019quantum,havlivcek2019supervised,lloyd1999quantum}. The existence of the quantum feature map hence the kernel in a quantum RKHS is guaranteed by the Theorems 1 \& 2 in~\cite{schuld2019quantum}.
 
There are a few strategies to implement a quantum kernel. In~\cite{schuld2019quantum,lloyd1999quantum}, the authors investigate kernels that can be classically efficiently computed (e.g. Gaussian) to facilitate conventional classifiers (i.e., SVM). \cite{havlivcek2019supervised,coyle2020born} provide a novel prospective to construct a feature map which is (conjectured to be) not classically computable efficiently. To be more concrete, for the case where the kernel is classically hard to estimate up to polynomial small error, computing it using a quantum system will only lead to an additive sampling error at most. Therefore, we adapt the design of quantum kernel from~\cite{havlivcek2019supervised,coyle2020born}:
\begin{gather}\label{eq:kernelcircuit}
    \Phi: \mathbf{x} \in \mathcal{X} \rightarrow \ket{\Phi(\mathbf{x})}\in\mathcal{H}_q,
\qquad
    \ket{\Phi(\mathbf{x})} =
    \boldsymbol{U}_{\Phi(\mathbf{x})}\boldsymbol H^{\otimes n}\boldsymbol{U}_{\Phi(\mathbf{x})}\boldsymbol H^{\otimes n}\ket{0}^{\otimes n} 
\end{gather}
For the favour of experiments, here we have also only considered single and two-qubit operations (Ising scenario), same as the majority system designs on the NISQ devices:
\begin{gather}\label{eq:qk_feat}
U_{\phi_{\{k\}}(\mathbf{x})}=\exp \left(\mathrm{i} \phi_{\{k\}}(\mathbf{x})\cdot Z_{k}\right) \qquad\quad\quad\text{(single qubit gate)}\\
U_{\phi_{\{l, m\}}(\mathbf{x})}=\exp \left(\mathrm{i} \phi_{\{l, m\}}(\mathbf{x})\cdot Z_{l} \otimes Z_{m}\right)\quad\text{(two-qubit gate)}
\end{gather}
where $Z_{l,m,k}$ are the $Z$-basis Pauli operator. The encode functions $\phi_{\{k\}}(\cdot)$ and $\phi_{\{l,m\}}(\cdot)$ can be chosen specifically to represent different classically-hard-to-learn feature maps\footnote{We recommend Figure.3b in~\cite{havlivcek2019supervised} for a visualisation example of a classically-hard-to-learn feature map.}:
\begin{gather}\label{eq:qk_encode}
    \phi_{\{k\}}(\mathbf{x})=\frac{\pi}{4} x_{k},
    \quad
    \phi_{\{l, m\}}(\mathbf{x})=\left(\frac{\pi}{4}-x_{l}\right)\left(\frac{\pi}{4}-x_{m}\right),
    \quad
    \text{from~\cite{coyle2020born}}\\
    \phi_{\{k\}}(\mathbf{x})=x_{k},
    \qquad
    \phi_{\{l, m\}}(\mathbf{x})=\left(\pi-x_{l}\right)\left(\pi-x_{m}\right),
    \qquad\,\,
    \text{from~\cite{havlivcek2019supervised}}
\end{gather}

Before providing implementation details, we first discuss the sample complexity of a quantum kernel in Theorem~\ref{theo:sample}. Then we show that this quantum kernel is bounded, measurable and characteristic in Theorem~\ref{theo:bound}, to prove its validity for the $\mathcal{L}_\mathrm{MMD}$-based training (in~\secref{subsec:loss}).

\begin{theorem}[Sample complexity for a quantum kernel and Gram matrix\label{theo:sample}]
To empirically estimate $\kappa_{q}(\mathbf{x}, \mathbf{y})=\langle\Phi(\mathbf{x}), \Phi(\mathbf{y})\rangle_{\mathcal{H}_{q}}$ up to a sampling error $\tilde{\epsilon}=\mathcal{O}(R^{-1/2})$, it requires $R=\mathcal{O}(\tilde{\epsilon}^2)$ times measurements on the quantum state: $\boldsymbol{U}_{\Phi(\mathbf{y})}^{\dagger} \boldsymbol{U}_{\Phi(\mathbf{x})}\ket{0}^{\otimes n}$. Assume the two sets of data have the same size of $N$: $N=\|\mathcal{X}\|=\|\mathcal{Y}\|$ where $\mathcal{X} = \{\mathbf{x}_1, \mathbf{x}_2,..., \mathbf{x}_N\}$ and $\mathcal{Y} = \{\mathbf{y}_1, \mathbf{y}_2,..., \mathbf{y}_N\}$. The sample complexity to estimate the Gram matrix $\left\{\mathcal{G}\mid\mathcal{G}_{ij}=\kappa_{q}(\mathbf{x}_i, \mathbf{y}_j)\right\}$ is expected to scale to $\mathcal{O}\left(\epsilon^{-2}N^{4}\right)$.
\end{theorem}
Note that the quantum states will collapse when the observation is made, therefore each measurement is made via restarting the system from the beginning. Moreover, Theorem~\ref{theo:sample} implies that the required number of samples will grow quadratically w.r.t. error and biquadratically w.r.t data size. This also makes an extreme challenge for the simulations with classical computation tools.

\begin{theorem}[Characteristic of a quantum kernel\label{theo:bound}]
Same as Gaussian kernel, the quantum kernel defined in~\eqref{eq:kernelcircuit} is also a characteristic/universal kernel. 
\end{theorem}
A requirement for MMD to be a valid metric to compare two distribution is that the kernel should be characteristic~\cite{sriperumbudur2011universality}, i.e.,  $\gamma_{\mathrm{MMD}}(\mathcal{P}, \mathcal{Q}) = 0\Longleftrightarrow \mathcal{P}\equiv\mathcal{Q}$, this is proven by Theorem~\ref{theo:bound}.
\newpage
Inspired by~\cite{havlivcek2019supervised,coyle2020born}, we propose two modes to encode the point set with quantum kernels. 
\begin{wrapfigure}[24]{r}{0.4\textwidth}
\vspace{6pt}
\centering
\includegraphics[width=1\linewidth]{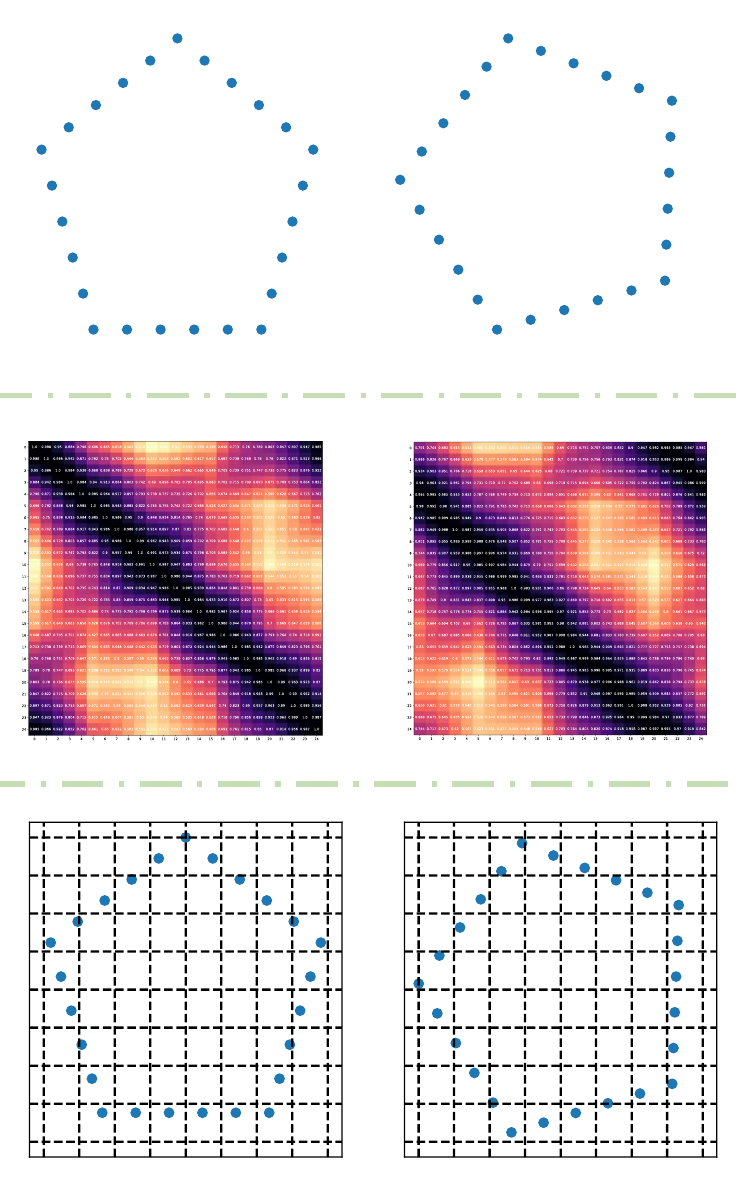}
\vspace{-8pt}
\renewcommand{\captionlabelfont}{\footnotesize}
\caption{\footnotesize Two implementations of $\kappa_q(\cdot, \cdot)$\label{fig:qkernel}}
\end{wrapfigure}
\paragraph{Continuous coordinates}
We can utilise the point coordinates for the $\mathbf{x}$ in~\eqref{eq:qk_encode}, then derive the Gram matrix after tremendous measurements~\cite{havlivcek2019supervised} (middle of~\figref{fig:qkernel}). It directly encode the continuous coordinates with no information loss when mapping from $\mathbb{R}^d$ to quantum RKHS. As shown in~\figref{fig:qkernel}, the derived Gram matrix represents some affinity measure for both all the point pairs within a single set (middle left) or between two sets (middle right). However, due the $Ising$ requirement at the implementation level, it is not straightforward to use $\phi_{\{l, m\}}(\mathbf{x})$ to encode data that are has more two dimensions\footnote{All the experiments in~\cite{havlivcek2019supervised} are based on a 2D dataset which consists of thirty data points from two categories.}. Therefore, we propose an alternative as follows.
\paragraph{Binned coordinates}
Inspired by~\cite{coyle2020born}, we first convert the each point into a 1D binary string by space quantization, then we substitute the quantized point coordinates for the $\mathbf{x}$ in~\eqref{eq:qk_encode} (bottom of~\figref{fig:qkernel}). In the right diagram, we use six qubits (three for each axis) to represent a pentagon. As a result, each point is represented via a 1D binary string whose length is six. This allows us to encode the 3D objects, therefore we use this design in our experiments in the next section. 

We provide more implementation details in the appendix.

\section{Experiments\label{sec:exp}}
In this section, we first describe the architectures and measurement procedures of qKC. We then present the set-up details and compare the experimental results with the prior attempt~\cite{golyanik2020quantum}.
\paragraph{Circuit architecture}
We use QAOA circuit~\cite{farhi2014quantum}, it is an algorithm to approximately prepare a desired quantum state via a $p$-depth quantum circuit. The quantum state encodes the solution for a \textit{specific} problem that can be extracted by measuring the final state after evolution. We use the same design of IBM~\cite{coyle2020born} (described in~\secref{subsec:qcbm}), to implement a shallowest version of a QAOA circuit (i.e., $p=1$) by choosing a particular set of $(\boldsymbol{\Gamma}, \boldsymbol{\Delta}, \boldsymbol{\Sigma})$:
\begin{gather}\label{eq:qaoaParam}
\QAOA_{p=1}(\{J_{ij}, b_{k}\}, \mathbf{\Gamma}) = \IBM\left(\boldsymbol\alpha=\{J_{ij}, b_{k}\}, \mathbf{\Gamma} = -\mathbf{\Gamma}, \mathbf{\Delta} = \mathbf{0}, \mathbf{\Sigma} = \mathbf{0}\right)
\end{gather}
The evolution defined by $
U_{f}^{k}(\mathbf{\Gamma}, \boldsymbol{\Delta}, \boldsymbol{\Sigma})
$ can be thus decomposed into a tensor product of single qubit unitaries, in corresponding to the rotations around the Pauli-$X$ axis:
\begin{gather}
U_{f}^{\QAOA}(\mathbf{\Gamma}, \boldsymbol{\Delta}, \mathbf{\Sigma}) = U_{f}\left(\forall k: \Gamma_{k}=-\Gamma_{k}, \Delta_{k}=0, \Sigma_{k}=0\right)=\exp \left(-i \sum_{k=1}^{n} \Gamma_{k} X_{k}\right)
\end{gather}
The output distribution (in~\eqref{eq:output_dist}) of the quantum circuit now becomes:
\begin{gather}\label{eq:output_qaoa}
    p^{\QAOA}_{\boldsymbol\alpha}(\mathbf{x}) = \left|\bra{\mathbf{x}}\boldsymbol U_{f}^{\QAOA} \prod_{j=1}^{m} U_{z}\left(\alpha_{j}, S_{j}\right)\boldsymbol{H}^{\otimes n}\ket{0}^{\otimes n}\right|^2
\end{gather}
where we optimise the circuit parameters $\boldsymbol\alpha=\{J_{ij}, b_{k}\}$ w.r.t objectives defined in~\eqref{eq:empirial_lmmd}. (Farhi \& Harrow~\cite{farhi2014quantum}) provide a thorough discussion on the (potential) supremacy of QAOA circuits when dealing with classically-hard-to-learn probability distributions.

We base our developments on the Rigetti Computing platform\footnote{Company webpage: \href{https://www.rigetti.com/}{https://www.rigetti.com/}, developer profile page: \href{https://github.com/rigetti}{https://github.com/rigetti/}}, which provides both simulation environment and API to the real quantum computers (QPUs).

\begin{figure}[t]
\centering
    \includegraphics[width=\linewidth]{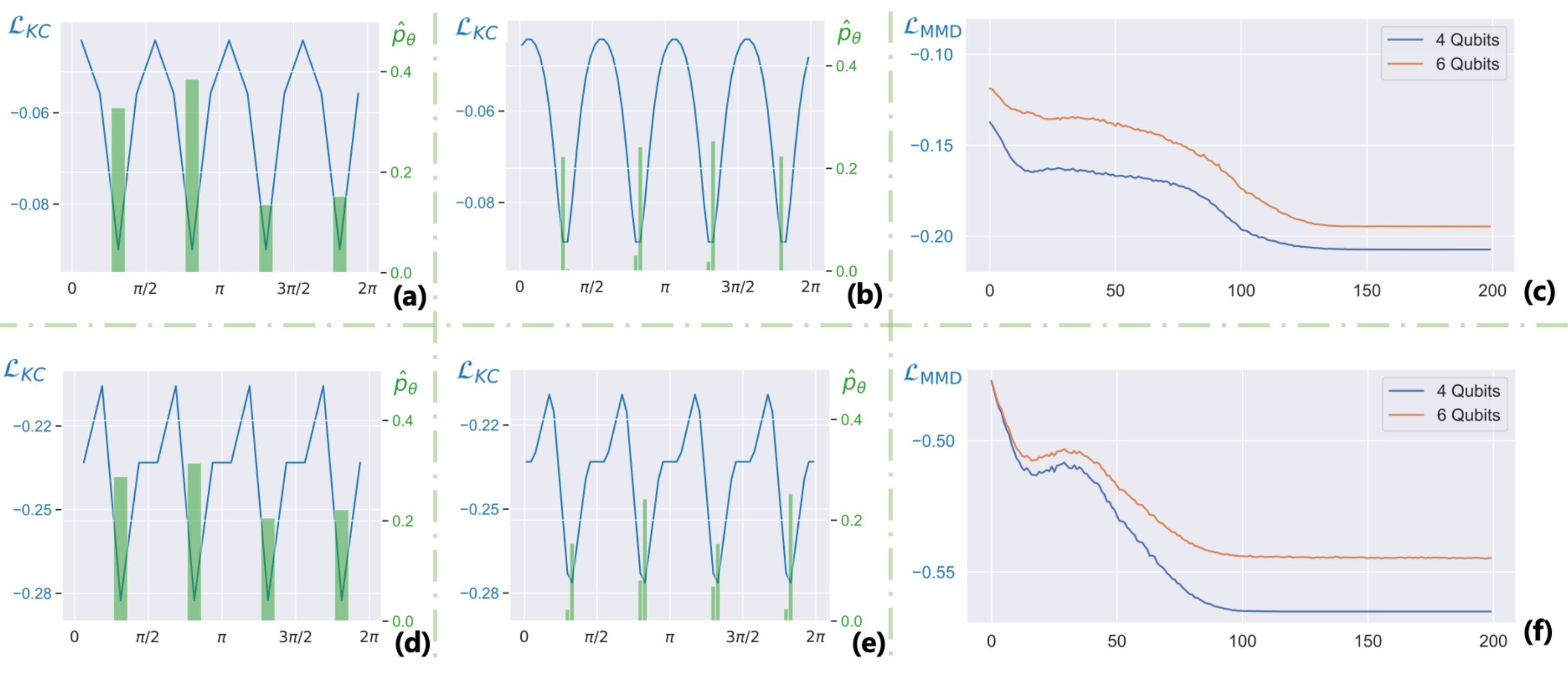}
    \caption{Matching a 2D (symmetric) square in the $xy$ plane with qKC, with a rotation of $5\pi/16$ around the $\hat{z}$-axis. (a) and (b): KC loss (in blue line) and output rotation distribution (in green bar) from the 4-qubit and 6-qubit qKC, respectively, (c): training curves of qKC. (d), (e) and (f) are based on the same setting, except using a quantum kernel instead of Gaussian kernels.\label{fig:training}} 
\end{figure}

\paragraph{Training setting}
Same as previous works~\cite{farhi2018classification,coyle2020born,golyanik2020quantum}, our experiments are limited by the fact that only small (in terms of number of operating qubits) quantum devices are simulatable and experimentally available. So we only stick to the 4-qubit and 6-qubit QAOA architectures. 

We use Adam optimiser, the learning rate is initially set to 0.02 and is decayed by 0.5 every 50 iterations. We set the mini-batch size as 1000 ($\|\mathcal{X}\|$ in~\eqref{eq:empirial_lmmd}). We find that the training optimisation highly depends on the design of the learning rate. Here we report the experimental results with optimal parameters, and we provide more ablation studies in appendix. We use zero-initialisation for the Ising parameters, $\{J_{ij}, b_{k}\}$, which ensures that the circuit outputs the uniform angle distribution. We highly encourage to take a look at the videos in the supplementary, for a qualitative understanding on how the output of the quantum circuit model evolve during the training.


\paragraph{Evaluation metric}
We use the same metrics as the prior attempt~\cite{golyanik2020quantum} to quantify the preciseness of different matching methods. The first metric named alignment error is defined as: 
$\|\mathbf{\mathcal{T}_{\hat{\theta}}\mathcal{M}-\mathcal{S} \|_{\mathcal{HS}}}/\|\mathbf{\mathcal{M}}\|_{\mathcal{HS}}$, 
where $\|\cdot\|_{\mathcal{HS}}$ denotes the Hilbert-Schmidt/Frobenius norm of a matrix. It measures how accurate the aligned shapes coincides with the ground truth transformation. We report its mean $e_{2d \text{ or } 3d}$ and standard derivation $\sigma_{2d \text{ or } 3d}$ over a sweep of angles between $0$ and $2\pi$. The second metric, transformation discrepancy, is defined as: $\|\mathbf{I-\mathcal{T}\mathcal{T}^\top}\|_{\mathcal{HS}}$. In contrast with~\cite{golyanik2020quantum}, qKC directly outputs the transformation angles. Therefore the mean $e_{\mathbf{R}}$ and std $\sigma_{\mathbf{R}}$ of this metric is always \textit{zero}. For asymmetric shapes, we choose angle with the highest probability as the optimal rotation.

\paragraph{2D symmetric shape}
We first evaluate qKC on the synthesised 2D symmetric polygons. Each polygon is centred at the origin and inscribed by a unit circle. We use the canonical and rotated shapes for the scene $\mathcal{S}$ and model $\mathcal{M}$ sets, respectively. Each side of these polygons consists of 10 points. In~\figref{fig:training}, we plot the KC loss, output distribution and training curve for matching a pair of squares, with a rotation of $5\pi/16$ (the other optima are $13\pi/16, 21\pi/16, 29\pi/16$). 

We show that both Gaussian kernel and quantum kernel loss can capture all the optimal rotations. For the 4-qubit qKC, all the output samples (1000 predictions) fall in the optimal solutions. As the resolution (i.e., number of qubits) increases, over 85\% of samples belong to the optima and 99\% are covered by their neighbourhoods. Compare with previous attempts~\cite{Diff_QBM,coyle2020born} on learning with Quantum Born Machine, our loss curve is with more stability while decreasing. We observe that the quantum kernel provides a faster convergence rate.

\paragraph{Fish}
\begin{table}[t]
\centering
\small
\begin{tabular}{l|ccccc|cccc}
    & \multicolumn{5}{c|}{Golyanik \& Theobalt, QA~\cite{golyanik2020quantum}} & \multicolumn{4}{c}{Ours, qKC} \\\toprule \bottomrule
    & \multicolumn{5}{c|}{$K$} & \multicolumn{2}{c}{$\kappa_G$} & \multicolumn{2}{c}{$\kappa_Q$} \\\hline
    & 10 & 20 & 30 & 40 & 50 & $n_Q=4$ & $n_Q=6$ & $n_Q=4$ & $n_Q=6$ \\\hline
    $e_{2D}$ & 0.026 & 0.041 & 0.078 & 0.17 & 0.3 & \textbf{0} & \textbf{0.0097} & \textbf{0} & \textbf{0.0127} \\
    $\sigma_{2D}$ & 0.013 & 0.012 & 0.012 & 0.012 & 0.013 & \textbf{0} & \textbf{0.0199} & \textbf{0} & \textbf{0.0084} \\
    $e_R$ & 0.062 & 0.083 & 0.22 & 0.47 & 0.764 & \textbf{0} & \textbf{0} & \textbf{0} & \textbf{0} \\
    $\sigma_R$ & 0.044 & 0.041 & 0.036 & 0.031 & 0.03 & \textbf{0} & \textbf{0} & \textbf{0} & \textbf{0} \\
    \end{tabular}
\vspace{1em}
\caption{Matching performance on the fish dataset (2D). We compare 4-qubit and 6-qubit ($n_Q$=4 or 6) qKC of Gaussian kernel ($\kappa_G, \sigma^2=0.01$) and quantum kernel ($\kappa_{Q}$), with all the configurations (characterised by the number of corresponding points, $K$) of QA.\label{tab:2d}}
\end{table}
We compare qKC with QA~\cite{golyanik2020quantum} on matching the 2D shapes from the fish dataset in Table~\ref{tab:2d}. Though the two methods are based on fundamentally different architectures, we stick to the same experimental settings as~\cite{golyanik2020quantum}. We use the rotation with the highest output probability for benchmarking. We have demonstrated that qKC can solve the matching problem with more preciseness.

\paragraph{Noise sensitivity}
\begin{wrapfigure}[8]{r}{0.5\textwidth}
\vspace{-6pt}
\begin{center}
\includegraphics[width=\linewidth]{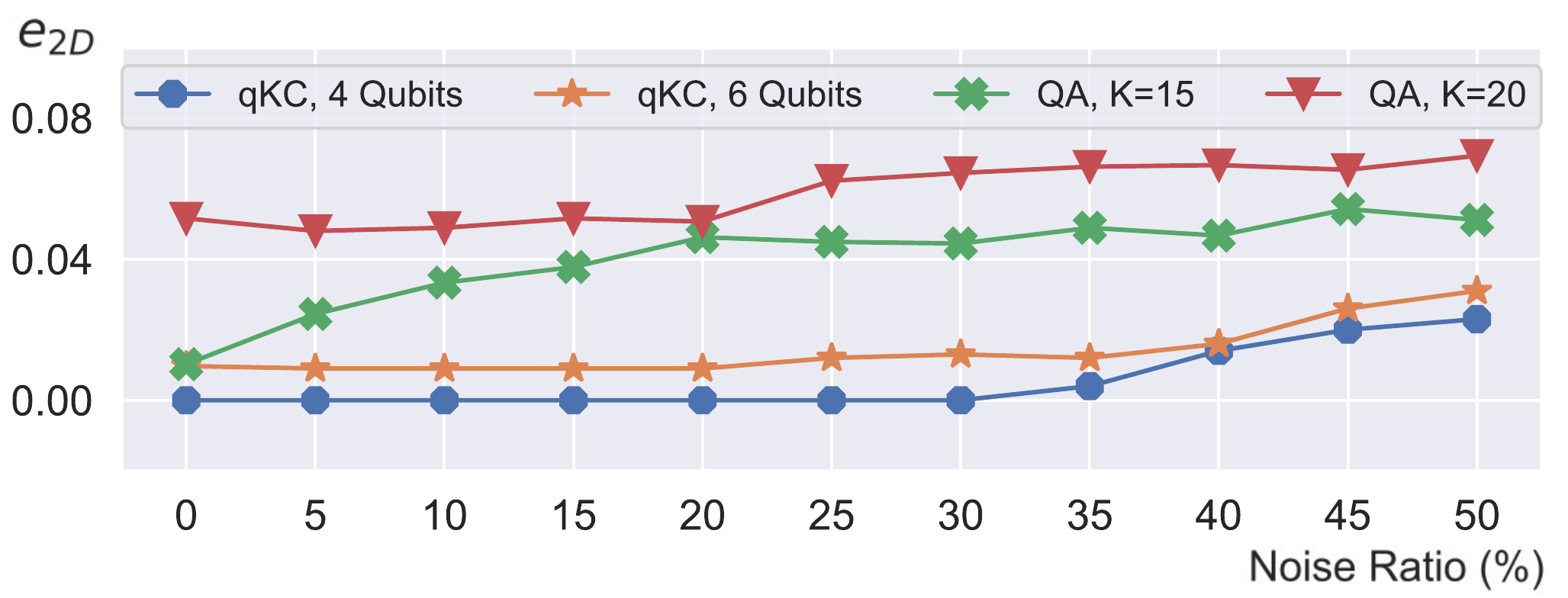}
\end{center}
\vspace{-12pt}
\renewcommand{\captionlabelfont}{\footnotesize}
\caption{\footnotesize Noise Sensitivity.\label{fig:noise}} 
\end{wrapfigure}
We use the same procedures as QA~\cite{golyanik2020quantum} to evaluate the robustness of qKC when the data (from fish dataset) is accompanied with noise. The ratio of iid Gaussian noise ranges from 5\% to 50\%. We report the averaged $e_{2D}$ for every noise ratio over 50 runs in~\figref{fig:noise}. We observe both methods have increasing misalignment when the noise level rises, and qKC has smaller generalisation error.

\paragraph{3D symmetric shape}
We manually select 3D symmetric shapes (e.g,~\figref{fig:diag_kc}) from ModelNet40 dataset~\cite{wu20153d}.
We provide more qualitative\footnote{Matching symmetric shapes are few investigated, for instance, in DCP~\cite{wang2019deep} and PoiotNetLK~\cite{aoki2019pointnetlk}, the range of rotations is $[0, \pi/4]$, and we do not find a proper benchmark metric. Since we didn't manage to hear from the authors of QA~\cite{golyanik2020quantum}, we don't have their implementations (not open-sourced) for a more thorough comparisons.} comparisons with conventional approaches incl. ICP~\cite{BeslM92, zhang1994iterative}, EM-ICP~\cite{EM-ICP} and deep learning methods incl. DCP~\cite{wang2019deep} and PoiotNetLK~\cite{aoki2019pointnetlk} in the appendix.

\section{Related Work}\label{sec:result}

Besides recent progress such as self supervised learning~\cite{pointcontrast,OcCo,zhang2021self} on 3D point cloud, quantum computation~\cite{qml_nature,PhysRevA.98.032309} provides a novel perspective for efficient processing and optimisation. While the previous attempts~\cite{golyanik2020quantum,benkner2020adiabatic,benkner2021q,birdal2021quantum} are based on the annealing-based quantum system (adiabatic quantum computer, AQC), we instead consider a gate-based quantum circuit learning framework.

To the best of our knowledge, we are the first to propose a QML approach for 3D vision tasks. The most relevant work is QA proposed by (Golyanik \& Theobalt~\cite{golyanik2020quantum}). QA formulates the matching objective as a quadratic unconstrained binary optimisation problem (QUBOP), where the measured qubit states are expected to provide solution for the pre-defined quadratic energy. However, their method is not empirically applicable to 3D data and cannot find multiple optimal rotations when the shapes are symmetric.
In contrast, we formulate this task as a distribution learning problem realised by a quantum circuit system. In comparison on the same downstream task, our method is more robust, and with less transformation discrepancy and alignment error. We have also provide a novel technique for quantum kernel encoding, which can cope with classically-hard-to-learn feature maps in RKHS.

\section{Discussion}\label{sec:discussion}

In this work, for the first time, a generic QCL framework is applied to the point set matching problem, whose solution is represented as a distribution of rotations. We have shown our proposed framework named qKC has several benefits in comparison with the prior work. We believe qKC is potentially capable of showing quantum computational advantage upon efficient implementation on quantum hardware in the near future.  
We hope that our work will bring insights for future research in quantum machine learning and related optimisation problems in the 3d vision community.

\newpage
\section*{Acknowledgement}
We would like to thank Weiyang Liu, Shuai Yuan, Clarice D. Aiello and Joan Lasenby for valuable discussions. HW is partially supported by the Cambridge Trust Scholarship, the Cathy Xu Fellowship, the CAPA Research Grant and the Cambridge Philosophical Society.

\bibliographystyle{splncs04}

\newpage

\clearpage
\newpage
\clearpage
\appendix


\section{Implementation Details of Quantum Kernels}
In~\secref{subsec:qkernel}, we propose two modes (continuous/binned coordinates) to encode a 2D/3D point set with quantum kernels, here we elaborate the implementation details and provide more discussions.

Instead of normalising objects into a unit sphere centred at the origin, here in both encoding procedures, we move the mass centres to $[0.5, 0.5, (0.5)]$ (also in a unit square (cubic)).

Essentially the unitaries of the quantum kernel encoding are very similar to that of the (Ising) Born Machine~\cite{coyle2020born} (in~\eqref{eq:evolution_unitary}), which can be written as:
\begin{gather}
    \boldsymbol{U}_\Phi(\mathbf{x})\coloneqq \prod_{j=1}^m \exp\left(\mathrm{i} \sum_{S_j \subseteq [n]}\phi_{S_j}(\mathbf{x}) \bigotimes_{k \in S_j} Z_k \right)
\end{gather}
where $\boldsymbol\theta (\boldsymbol\alpha)$ in~\eqref{eq:evolution_unitary} here is replaced by the feature map $\phi(\mathbf{x})$ with sample $\mathbf{x}$.

Clearly from the design of $\phi(\mathbf{x})$ in~\eqref{eq:qk_feat}, we only consider the cases of $n=2$ (i.e., Ising). The circuit\footnote{also is it very close to IQP circuit class} implementation to calculate quantum kernel is also very similar to the ones used in qKC.
\begin{wrapfigure}[10]{r}{0.5\textwidth}
\centering
\includegraphics[width=1\linewidth]{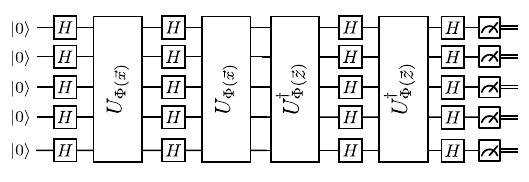}
\vspace{-16pt}
\renewcommand{\captionlabelfont}{\footnotesize}
\caption{\footnotesize Quantum Kernel Calculation, from~\cite{havlivcek2019supervised}\label{fig:qkernels1}}
\end{wrapfigure}
Same as IBM model, the inner product based on quantum kernel can be empirically estimated from a set of measurements, or, it can be \textit{precisely} derived from the direct calculation on the overlapping of two wave functions. Note that the latter is only applicable in simulations, and the previous one can used in both simulated and real circuits. In all our experiments (simulation-based), we use the \textit{exact calculation} instead of estimating from measurements. But it is worth discussing the sample complexity of the quantum kernel (correlation), which we will investigate in next section. Before that, here we will describe how to construct the (empirical) estimator for the quantum kernel inner product:
\begin{gather}\label{eq:qkernel_supple}
\kappa_q(\mathbf{x}, \mathbf{y})
= \langle\Phi(\mathbf{x}), \Phi(\mathbf{y})\rangle_{\mathcal{H}_q} 
= |\braket{\Phi(\mathbf{x})\,|\,\Phi(\mathbf{y})}|^2
\end{gather}

The conventional method of estimating the fidelity (i.e., overlap) between two states is the SWAP test~\cite{buhrman2001quantum}. However, it is not feasible on NISQ devices since it can be realised on neither shallow-depth circuits nor sub-universal gates. Therefore we use the recent developed protocols proposed in~\cite{cincio2018learning,havlivcek2019supervised}. In fact, the form of~\eqref{eq:qkernel_supple} already indicate us how to design the measure procedure. We can use the circuit shown in~\figref{fig:qkernels1}, simply apply the circuit $\boldsymbol{U}_{\Phi(\vec{y})}^{\dagger} \boldsymbol{U}_{\Phi(\vec{x})}$ to the state $|0\rangle^{\otimes n}$. The sample (repeat measurements) $R$ times the resulting state in the $Z$-basis. Record the number of all observed zero bit strings, and then divide by the total number of shots $R$, which will give us a estimator for $\kappa_q(\mathbf{x}, \mathbf{y})$:
\begin{gather}
    \hat{\kappa}(\mathbf{x}, \mathbf{y}) = \#\,\,\text{of }\{0,0,...,0\}\cdot R^{-1}
\end{gather}



Empirically we found that training qKC with quantum kernels seems to have faster convergence, as reported in the previous study~\cite{coyle2020born}. However, computation of quantum kernels is likely not to be feasible with NISQ devices. 
In our case, it takes \textbf{more than a week} to empirically calculate a sweep of KC (64 float values for a 6-qubit qKC, shown in (e) in~\figref{fig:training}) for aligning two 2D symmetric square (10 points on each side, 36 points in total)\footnote{this is also reported in~\cite{coyle2020born,havlivcek2019supervised}. Therefore, the majority of our experiments is based on Gaussian kernels.}. Furthermore, current calculation procedure~\cite{coyle2020born,havlivcek2019supervised} has the "curse of scalability". Roughly speaking, for the case of qKC, the total evaluation time increases in a biquadrate rate w.r.t. the number of points in a set, and in a exponential rate w.r.t. the number of qubits of IBM. As a comparison, it takes \textit{at most} a few minutes to carry out the same evaluation procedure with classical Gaussian kernels on a home computer.

\textbf{We have provided comprehensive comments/docstrings/instructions in the scripts for quantum kernel calculations. We highly recommend to read them if there are further questions.}

\section{Proof of Theorem. 2}
We use the same analogy as~\cite{havlivcek2019supervised}, and a more thorough analysis is provided in~\cite{Tropp15}.

We denote the ground true Gram matrix and its empirical estimation as $K$ and $\hat{K}$, respectively. A rough estimation on the operator norm $\|\cdot\|$ is upper bounded by the Frobenius norm $\|\cdot\|_F$:
\begin{gather}
    \|K-\hat{K}\| \leq \|K-\hat{K}\|_F = \sqrt{\sum_{i,j}\left|K_{ij}-\hat{K}_{ij}\right|^2}\leq \Tilde{\epsilon}N
\end{gather}
where $\Tilde{\epsilon}$ is the largest sampling error of all the entries, $N$ is the number of data points: $N=\|\mathcal{X}\|=\|\mathcal{Y}\|$ ($\mathcal{X} = \{\mathbf{x}_1, \mathbf{x}_2,..., \mathbf{x}_N\}$ and $\mathcal{Y} = \{\mathbf{y}_1, \mathbf{y}_2,..., \mathbf{y}_N\}$).

Hence to ensure the maximum deviation of $\epsilon$ with high probability, each entry requires a number of $R=\mathcal{O}(\epsilon^{-2}|N|^2)$ measurements. For a total of $N(N-1)/2$ unique entries in $\hat{K}$, the full sampling complexity is expected to have a scale of $\mathcal{O}(\epsilon^{-2}N^4)$. The biquadrate rate w.r.t. $N$ for estimating the Gram matrices is also a challenge for efficiently implementing quantum kernels on current NISQ devices\footnote{as shown in the Fig.4 and Fig.S7 of~\cite{havlivcek2019supervised}, there are certain amount of entries with large gaps between experimental measurements (denoted as 'exp' in the figure legend) and theoretical calculations (same, as 'ideal').}.

\section{Tuning the Training Parameters}

There are not much hyper-parameters need to be tuned in the training process of qKC. Empirically we have noticed that the design of learning rate and batch size will affect the training process of qKC. For the Adam optimiser, we use the default configuration: 1) the coefficients used for computing running mean of gradient and its square are set as 0.9 and 0.999, respectively, 2) $eps$ is set as 1e-8 for numerical stability and 3) no L2 penalty term.

\begin{figure}[H]
\centering
    \includegraphics[width=\linewidth]{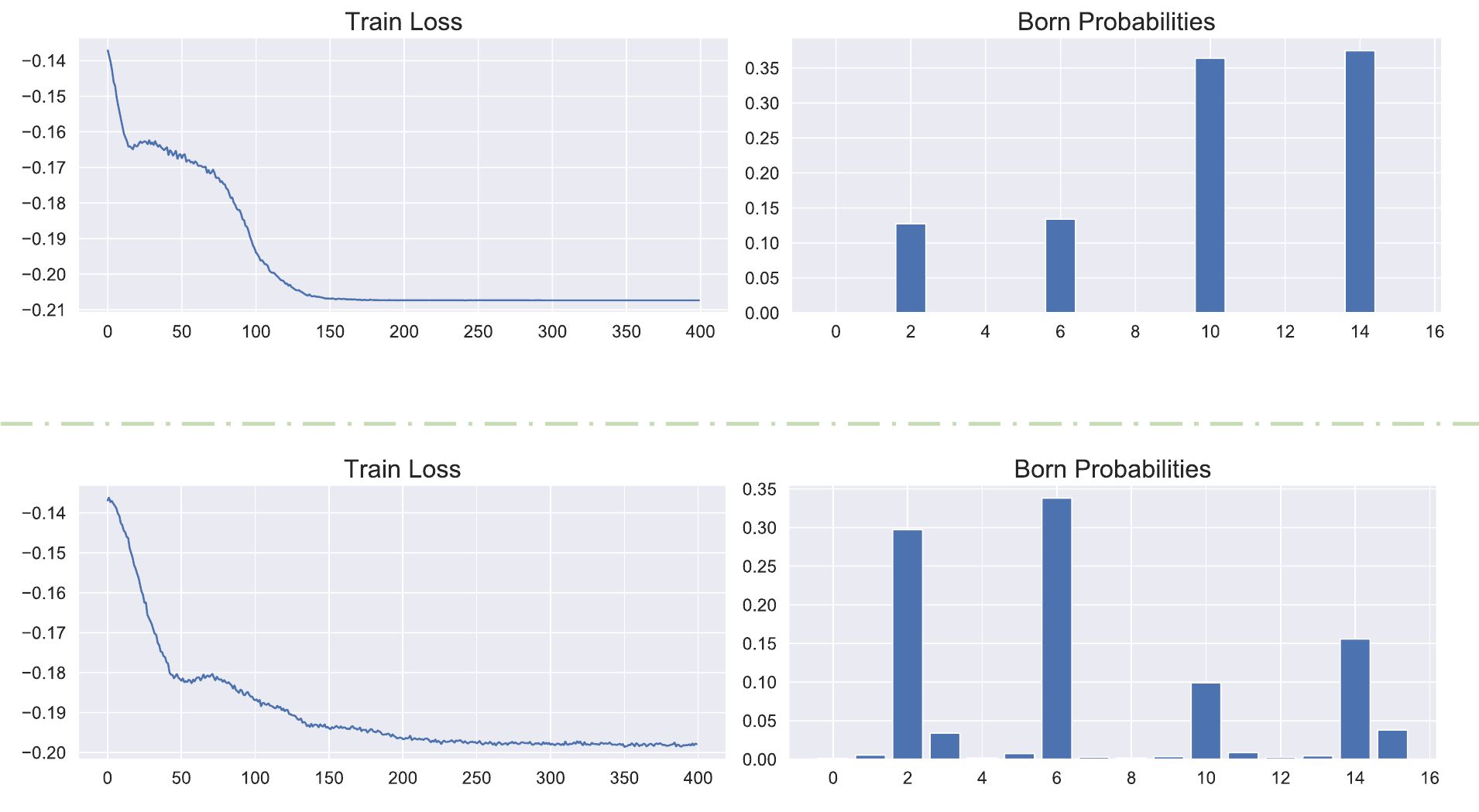}
    \caption{Batch size in above: 100; below: 1. In experiments we use a batch size of 1000.\label{fig:ablation_batch}} 
\end{figure}

In~\figref{fig:ablation_batch}, we change the batch size for the MMD estimator, while other configurations in the inset (a) of~\figref{fig:training} remain unchanged. It is clear that benefiting from the quadratic convergence rate of the unbiased MMD estimator (in~\eqref{eq:mmd_convergence}), a batch size of 100 is sufficient for the training iteration.

\clearpage
\newpage
On the other side, the design of learning rate is \emph{critical} for the qKC training. 

\begin{figure}[t]
\centering
    \includegraphics[width=\linewidth]{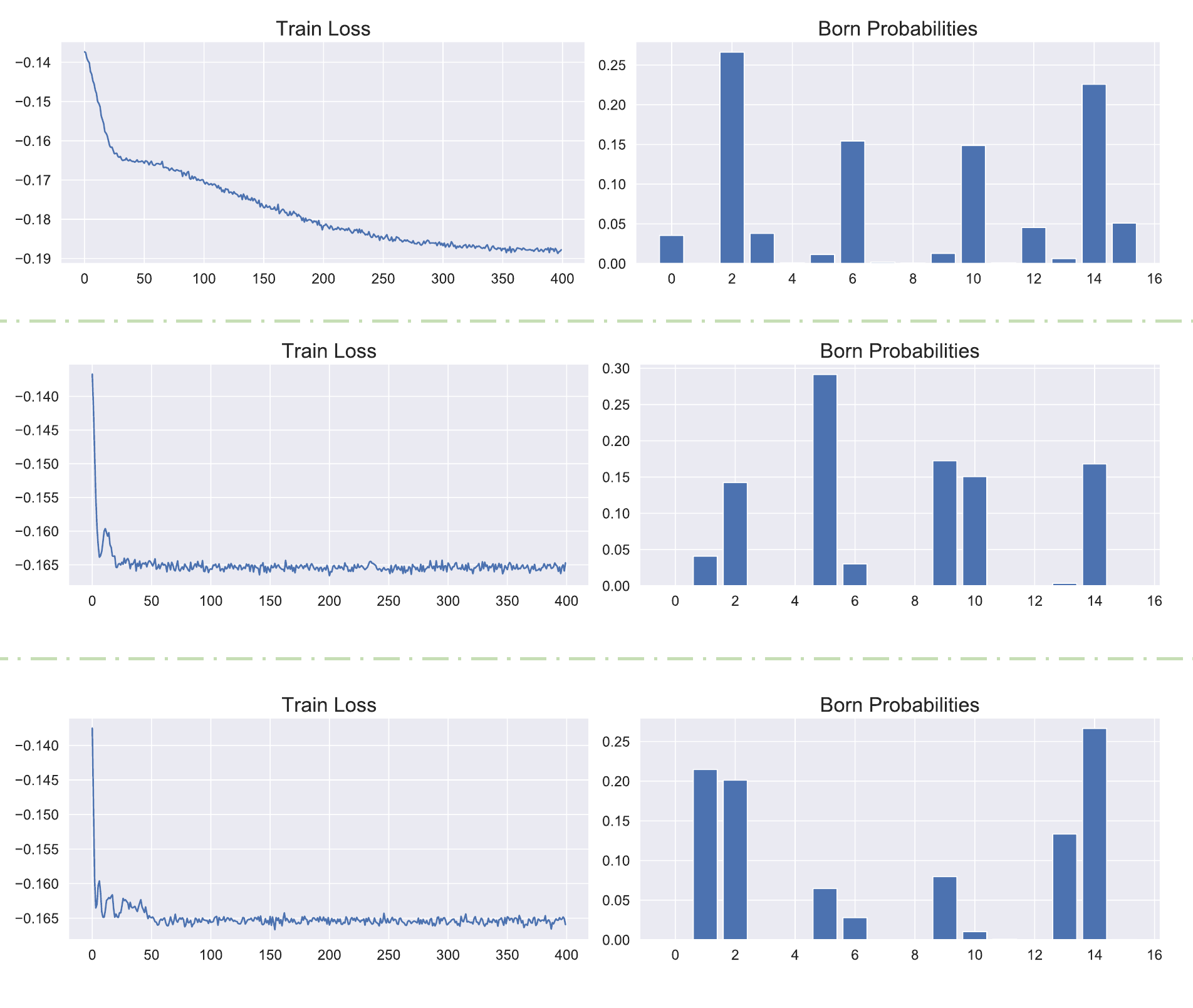}
    \caption{LR Init in upper: 0.01; middle: 0.05; bottom: 0.1. Experiments use an initial value of 0.02.\label{fig:ablation_lr_init}} 
\end{figure}

We first vary the staring point of the learning rate, where all are still with half decay every 50 epochs, as shown in~\figref{fig:ablation_lr_init}. From the empirical results, we found that a smaller initial value of learning rate 0.01() still can effectively decrease the loss but require much longer iteration steps (in comparison with inset (a) of~\figref{fig:training}). And a larger starting point (e.g., 0.05 and 0.1) will have a much steeper decreasing slope at the beginning but is hard to be optimised.

We omit the exploration on the decay design of the learning rate here, we find a half decay every 50 epochs is good enough for the optimisation of qKC, after trying with different step length and rate of the step decay, as well as others including cosine annealing and exponential decay.

\section{Qualitative Comparisons with Conventional and Deep Methods}

Here we mainly compare qKC with ICP, EM-ICP, PointNet-LK and DCP on matching 3D symmetric shapes (shown in the top-right inset of~\figref{fig:diag_kc}). 

We found that, with different starting points (i.e., initial point pair correspondence), methods like ICP and EM-ICP can also recover all the optimal angles, and are quite robust (i.e., not trapped by spurious local minima) to simple shapes such as chairs, tables and bookcases.

For the deep learning based methods, since the neural models are trained in a supervised fashion (single label $y$ for each input data point $\mathbf{x}$), it is not straightforward to design an output head that is capable of covering multiple optimal solutions jointly. Therefore, the authors constrained the rotated angles in a narrow range of $[0, \pi/4]$, which will cause trouble if the ground-true rotation is outside. We directly use the trained model weights provided by the authors and conduct the inference on these symmetric shapes for 100 times, and shown in the below figures.

\begin{figure}[t]
\centering
    \includegraphics[width=\linewidth]{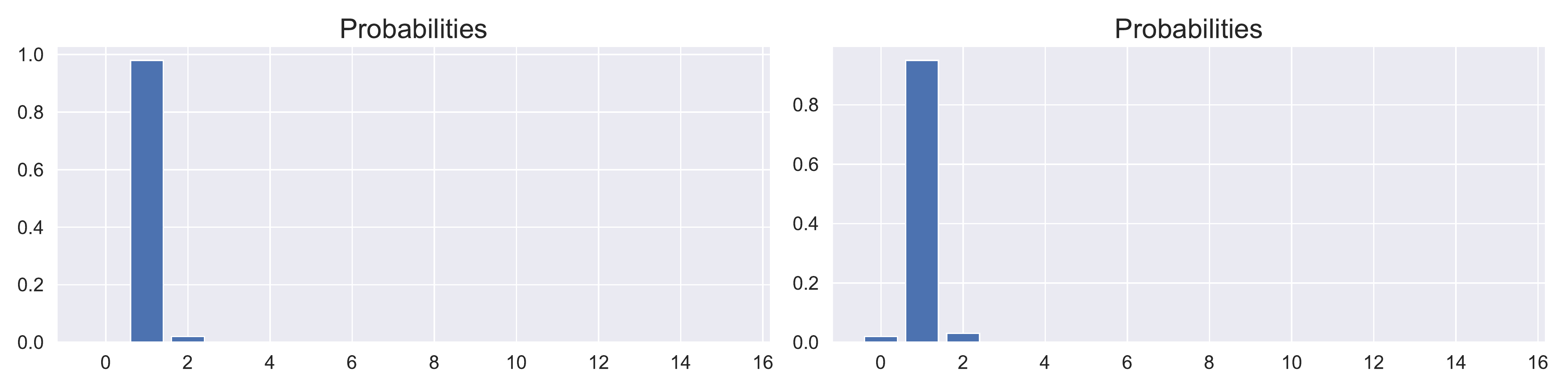}
    \caption{Left: DCP, Right: PointNet-LK\label{fig:3ddl}} 
\end{figure}

\begin{figure}[t]
\centering
    \includegraphics[width=\linewidth]{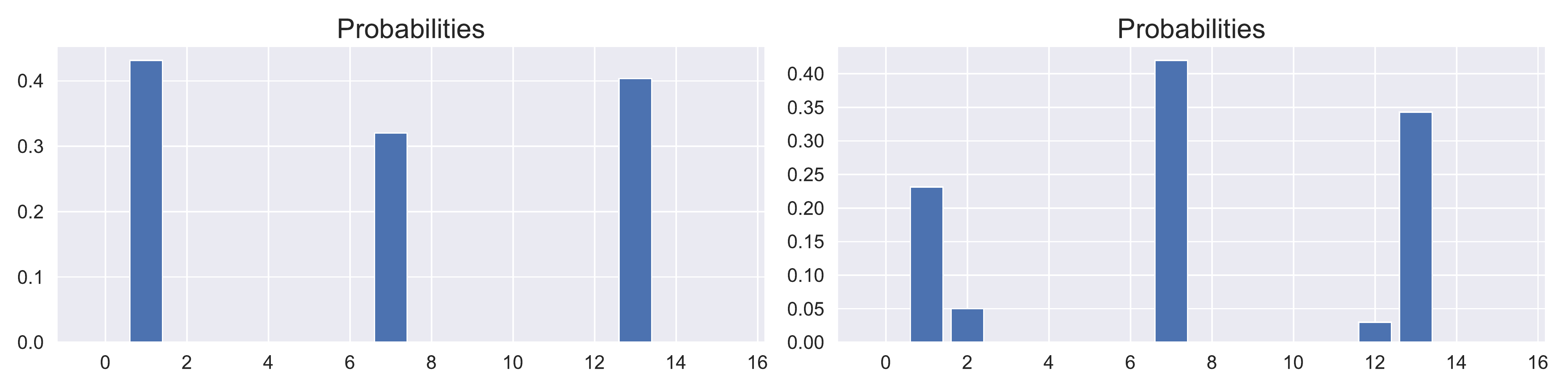}
    \caption{Left: qKC, Right: ICP\label{fig:3dqkc}} 
\end{figure}

\end{document}